%% file: paper.tex
\documentclass[10pt,twocolumn,letterpaper]{article}

\usepackage[pagenumbers]{cvpr} 

\usepackage{graphicx}
\usepackage{amsmath}
\usepackage{amssymb}
\usepackage{booktabs}
\usepackage{floatrow}
\usepackage{float}
\usepackage{multicol,graphicx,mathrsfs,pifont,amscd,latexsym,color,url, pifont, epstopdf,setspace,multirow,colortbl, bbm,listings, balance,color,indentfirst,enumitem, amsmath,amssymb, verbatim,microtype,xcolor}
\usepackage{graphicx,booktabs, paralist}

\usepackage[pagebackref,breaklinks,colorlinks]{hyperref}
\usepackage{fontawesome}

\input{plots_cvpr}

\usepackage[pagebackref,breaklinks,colorlinks]{hyperref}

\usepackage[capitalize]{cleveref}
\crefname{section}{Sec.}{Secs.}
\Crefname{section}{Section}{Sections}
\Crefname{table}{Table}{Tables}
\crefname{table}{Tab.}{Tabs.}

\def\cvprPaperID{4403} 
\def\confName{CVPR}
\def\confYear{2023}

\newcommand{\citep}{\cite}
\newcommand{\citet}{\cite}

\begin{document}

\input{abbrev}
\newcommand{\exstd}[1]{{\tiny\textit{}{$\pm$#1}}}   

\title{Improving Image Recognition by Retrieving from Web-Scale Image-Text Data}

\author{
Ahmet Iscen \ \ \ \ Alireza Fathi\ \ \ \ Cordelia Schmid\\
{\fontsize{11}{13}\selectfont Google Research}\\
}

\maketitle

\input{abstract}
\input{intro}

\input{related}

\input{method}

\input{memory}

\input{experiments}
\input{conclusions}

\newpage

{\small
\bibliographystyle{ieee_fullname}
\bibliography{egbib}
}
\clearpage
\newpage

\input{appendix}

\end{document}

%% file: plots_cvpr.tex
\makeatletter
\@namedef{ver@everyshi.sty}{}
\makeatother
\usepackage{tikz}

\usetikzlibrary{arrows,shapes,calc,matrix,fit,backgrounds}
\usepackage{pgfplots}
\usepackage{pgfplotstable}
\pgfplotsset{compat=1.9}

\usepackage{xstring}

\usepgfplotslibrary{external}

\newcommand{\extdata}[1]{\input{#1}}
\IfBeginWith*{\jobname}{fig/extern/}{\finalcopy}{}

\newcommand{\leg}[1]{\addlegendentry{#1}}

\tikzset{every mark/.append style={solid}}
\pgfplotsset{
	grid=both, width=\columnwidth, try min ticks=5,
	every axis/.append style={font=\scriptsize},
	every axis plot/.append style={thick,mark=none,mark size=1.2,tension=0.18},
	legend cell align=left, legend style={fill opacity=0.8},
}

\pgfplotsset{
	dash/.style={mark=o,dashed,opacity=0.7},
	dott/.style={mark=o,dotted,opacity=0.7},
}

%% file: abbrev.tex
\newcommand{\nn}[1]{\ensuremath{\text{NN}_{#1}}\xspace}

\newcommand{\citemiss}{\alert{[??]}\xspace}

\newcommand{\supe}[1]{^{\mkern-2mu(#1)}}
\newcommand{\dime}[1]{(#1)}

\def\l1{\ensuremath{\ell_1}\xspace}
\def\l2{\ensuremath{\ell_2}\xspace}

\newcommand*\OK{\ding{51}}

\newenvironment{narrow}[1][1pt]
	{\setlength{\tabcolsep}{#1}}
	{\setlength{\tabcolsep}{6pt}}

\newcommand{\commentout}[1]{}
\newcommand{\prm}[1]{_{#1}}

\newcommand{\alert}[1]{{\color{red}{#1}}}
\newcommand{\head}[1]{{\noindent\bf #1}}  
\newcommand{\equ}[1]{(\ref{equ:#1})\xspace}

\newcommand{\red}[1]{{\color{red}{#1}}}
\newcommand{\blue}[1]{{\color{blue}{#1}}}
\newcommand{\green}[1]{{\color{green}{#1}}}
\newcommand{\gray}[1]{{\color{gray}{#1}}}


\newcommand{\tran}{^\top}
\newcommand{\mtran}{^{-\top}}
\newcommand{\zcol}{\mathbf{0}}
\newcommand{\zrow}{\zcol\tran}

\newcommand{\ind}{\mathbbm{1}}
\newcommand{\expect}{\mathbb{E}}
\newcommand{\nat}{\mathbb{N}}
\newcommand{\zahl}{\mathbb{Z}}
\newcommand{\real}{\mathbb{R}}
\newcommand{\proj}{\mathbb{P}}
\newcommand{\prob}{\mathbf{Pr}}

\newcommand{\mif}{\textrm{if }}
\newcommand{\other}{\textrm{otherwise}}
\newcommand{\minimize}{\textrm{minimize }}
\newcommand{\maximize}{\textrm{maximize }}
\newcommand{\st}{\textrm{subject to }}

\newcommand{\id}{\operatorname{id}}
\newcommand{\const}{\operatorname{const}}
\newcommand{\sgn}{\operatorname{sgn}}
\newcommand{\var}{\operatorname{Var}}
\newcommand{\mean}{\operatorname{mean}}
\newcommand{\trace}{\operatorname{tr}}
\newcommand{\diag}{\operatorname{diag}}
\newcommand{\vect}{\operatorname{vec}}
\newcommand{\cov}{\operatorname{cov}}

\newcommand{\softmax}{\operatorname{softmax}}
\newcommand{\clip}{\operatorname{clip}}

\newcommand{\defn}{\mathrel{:=}}
\newcommand{\peq}{\mathrel{+\!=}}
\newcommand{\meq}{\mathrel{-\!=}}

\newcommand{\floor}[1]{\left\lfloor{#1}\right\rfloor}
\newcommand{\ceil}[1]{\left\lceil{#1}\right\rceil}
\newcommand{\inner}[1]{\left\langle{#1}\right\rangle}
\newcommand{\norm}[1]{\left\|{#1}\right\|}
\newcommand{\frob}[1]{\norm{#1}_F}
\newcommand{\card}[1]{\left|{#1}\right|\xspace}
\newcommand{\diff}{\mathrm{d}}
\newcommand{\der}[3][]{\frac{d^{#1}#2}{d#3^{#1}}}
\newcommand{\pder}[3][]{\frac{\partial^{#1}{#2}}{\partial{#3^{#1}}}}
\newcommand{\ipder}[3][]{\partial^{#1}{#2}/\partial{#3^{#1}}}
\newcommand{\dder}[3]{\frac{\partial^2{#1}}{\partial{#2}\partial{#3}}}

\newcommand{\wb}[1]{\overline{#1}}
\newcommand{\wt}[1]{\widetilde{#1}}

\def\xxssp{\hspace{-3pt}}
\def\xssp{\hspace{1pt}}
\def\ssp{\hspace{3pt}}
\def\msp{\hspace{5pt}}
\def\lsp{\hspace{12pt}}

\newcommand{\cA}{\mathcal{A}}
\newcommand{\cB}{\mathcal{B}}
\newcommand{\cC}{\mathcal{C}}
\newcommand{\cD}{\mathcal{D}}
\newcommand{\cE}{\mathcal{E}}
\newcommand{\cF}{\mathcal{F}}
\newcommand{\cG}{\mathcal{G}}
\newcommand{\cH}{\mathcal{H}}
\newcommand{\cI}{\mathcal{I}}
\newcommand{\cJ}{\mathcal{J}}
\newcommand{\cK}{\mathcal{K}}
\newcommand{\cL}{\mathcal{L}}
\newcommand{\cM}{\mathcal{M}}
\newcommand{\cN}{\mathcal{N}}
\newcommand{\cO}{\mathcal{O}}
\newcommand{\cP}{\mathcal{P}}
\newcommand{\cQ}{\mathcal{Q}}
\newcommand{\cR}{\mathcal{R}}
\newcommand{\cS}{\mathcal{S}}
\newcommand{\cT}{\mathcal{T}}
\newcommand{\cU}{\mathcal{U}}
\newcommand{\cV}{\mathcal{V}}
\newcommand{\cW}{\mathcal{W}}
\newcommand{\cX}{\mathcal{X}}
\newcommand{\cY}{\mathcal{Y}}
\newcommand{\cZ}{\mathcal{Z}}

\newcommand{\vA}{\mathbf{A}}
\newcommand{\vB}{\mathbf{B}}
\newcommand{\vC}{\mathbf{C}}
\newcommand{\vD}{\mathbf{D}}
\newcommand{\vE}{\mathbf{E}}
\newcommand{\vF}{\mathbf{F}}
\newcommand{\vG}{\mathbf{G}}
\newcommand{\vH}{\mathbf{H}}
\newcommand{\vI}{\mathbf{I}}
\newcommand{\vJ}{\mathbf{J}}
\newcommand{\vK}{\mathbf{K}}
\newcommand{\vL}{\mathbf{L}}
\newcommand{\vM}{\mathbf{M}}
\newcommand{\vN}{\mathbf{N}}
\newcommand{\vO}{\mathbf{O}}
\newcommand{\vP}{\mathbf{P}}
\newcommand{\vQ}{\mathbf{Q}}
\newcommand{\vR}{\mathbf{R}}
\newcommand{\vS}{\mathbf{S}}
\newcommand{\vT}{\mathbf{T}}
\newcommand{\vU}{\mathbf{U}}
\newcommand{\vV}{\mathbf{V}}
\newcommand{\vW}{\mathbf{W}}
\newcommand{\vX}{\mathbf{X}}
\newcommand{\vY}{\mathbf{Y}}
\newcommand{\vZ}{\mathbf{Z}}

\newcommand{\va}{\mathbf{a}}
\newcommand{\vb}{\mathbf{b}}
\newcommand{\vc}{\mathbf{c}}
\newcommand{\vd}{\mathbf{d}}
\newcommand{\ve}{\mathbf{e}}
\newcommand{\vf}{\mathbf{f}}
\newcommand{\vg}{\mathbf{g}}
\newcommand{\vh}{\mathbf{h}}
\newcommand{\vi}{\mathbf{i}}
\newcommand{\vj}{\mathbf{j}}
\newcommand{\vk}{\mathbf{k}}
\newcommand{\vl}{\mathbf{l}}
\newcommand{\vm}{\mathbf{m}}
\newcommand{\vn}{\mathbf{n}}
\newcommand{\vo}{\mathbf{o}}
\newcommand{\vp}{\mathbf{p}}
\newcommand{\vq}{\mathbf{q}}
\newcommand{\vr}{\mathbf{r}}
\newcommand{\vt}{\mathbf{t}}
\newcommand{\vu}{\mathbf{u}}
\newcommand{\vv}{\mathbf{v}}
\newcommand{\vw}{\mathbf{w}}
\newcommand{\vx}{\mathbf{x}}
\newcommand{\vy}{\mathbf{y}}
\newcommand{\vz}{\mathbf{z}}

\newcommand{\vone}{\mathbf{1}}
\newcommand{\vzero}{\mathbf{0}}

\newcommand{\valpha}{{\boldsymbol{\alpha}}}
\newcommand{\vbeta}{{\boldsymbol{\beta}}}
\newcommand{\vgamma}{{\boldsymbol{\gamma}}}
\newcommand{\vdelta}{{\boldsymbol{\delta}}}
\newcommand{\vepsilon}{{\boldsymbol{\epsilon}}}
\newcommand{\vzeta}{{\boldsymbol{\zeta}}}
\newcommand{\veta}{{\boldsymbol{\eta}}}
\newcommand{\vtheta}{{\boldsymbol{\theta}}}
\newcommand{\viota}{{\boldsymbol{\iota}}}
\newcommand{\vkappa}{{\boldsymbol{\kappa}}}
\newcommand{\vlambda}{{\boldsymbol{\lambda}}}
\newcommand{\vmu}{{\boldsymbol{\mu}}}
\newcommand{\vnu}{{\boldsymbol{\nu}}}
\newcommand{\vxi}{{\boldsymbol{\xi}}}
\newcommand{\vomikron}{{\boldsymbol{\omikron}}}
\newcommand{\vpi}{{\boldsymbol{\pi}}}
\newcommand{\vrho}{{\boldsymbol{\rho}}}
\newcommand{\vsigma}{{\boldsymbol{\sigma}}}
\newcommand{\vtau}{{\boldsymbol{\tau}}}
\newcommand{\vupsilon}{{\boldsymbol{\upsilon}}}
\newcommand{\vphi}{{\boldsymbol{\phi}}}
\newcommand{\vchi}{{\boldsymbol{\chi}}}
\newcommand{\vpsi}{{\boldsymbol{\psi}}}
\newcommand{\vomega}{{\boldsymbol{\omega}}}

\newcommand{\rLambda}{\mathrm{\Lambda}}
\newcommand{\rSigma}{\mathrm{\Sigma}}

\newcommand{\loss}{\mathcal{L}}

\def\onedot{.\xspace}
\def\eg{\emph{e.g}\onedot} \def\Eg{\emph{E.g}\onedot}
\def\ie{\emph{i.e}\onedot} \def\Ie{\emph{I.e}\onedot}
\def\cf{\emph{cf}\onedot} \def\Cf{\emph{C.f}\onedot}
\def\etc{\emph{etc}\onedot}
\def\vs{\emph{vs}\onedot}
\def\wrt{w.r.t\onedot} \def\dof{d.o.f\onedot}
\def\etal{\emph{et al}\onedot}

\newcommand{\std}[1]{{\tiny\textpm{}{#1}}}

\makeatother

%% file: abstract.tex
\begin{abstract}

Retrieval augmented models are becoming increasingly popular for computer vision tasks after their recent success in NLP problems.
The goal is to enhance the recognition capabilities of the model by retrieving similar examples for the visual input from an external memory set.
In this work, we introduce an attention-based memory module, which learns the importance of each retrieved example from the memory.
Compared to existing approaches, our method removes the influence of the irrelevant retrieved examples, and retains those that are beneficial to the input query.
We also thoroughly study various ways of constructing the memory dataset.
Our experiments show the benefit of using a massive-scale memory dataset of $1$B image-text pairs, and demonstrate the performance of different memory representations.
We evaluate our method in three different classification tasks, namely long-tailed recognition, learning with noisy labels, and fine-grained classification, and show that it achieves state-of-the-art accuracies in ImageNet-LT, Places-LT and Webvision datasets. 

\end{abstract}

%% file: intro.tex
\section{Introduction}

Increasing the number of parameters of large transformer models has been a recent successful trend achieving new benchmarks in vision and language tasks. Recent results from  T5~\cite{DBLP:journals/jmlr/RaffelSRLNMZLL20}, GPT-3~\cite{DBLP:conf/nips/BrownMRSKDNSSAA20}, PaLM~\cite{DBLP:journals/corr/abs-2204-02311},  CoCa~\cite{DBLP:journals/corr/abs-2205-01917}, Flamingo~\cite{DBLP:journals/corr/abs-2204-14198}, BEIT-3~\cite{DBLP:journals/corr/abs-2208-10442}, PaLI~\cite{pali2022}, Florence~\cite{yuan2021florence} and FLAVA~\cite{singh2022flava} show that transformer models are able to store a surprising amount of information when scaled to tens of billions of parameters and trained on vast text and image corpora. These so-called `foundation models' achieve state-of-the-art results when fine tuned and applied to secondary tasks such as language modeling, image captioning, visual question answering and open vocabulary recognition.

In these foundation models, the learned world knowledge is stored implicitly in the parameters of the underlying neural network. 
This implies that some of the problems of the current ML paradigm are amplified in these models: (a)~scaling is challenging, both in learning and serving, given the large number of parameters that are required for storing the knowledge, (b)~it is hard to update the model as the world facts change or input data gets modified, (c)~these models tend to be black box, which means it is hard to interpret the underlying reason behind their decisions.

To address the above issues, we propose an alternative perspective on the problem. 
Instead of compiling the world knowledge statically into model weights, we take an interpretive view where the world knowledge gets transformed into a massive-scale index/memory.
On the other hand, a relatively low-compute small model learns to use the memory for the given inference task. Instead of increasing the size of the model and training on more data as done in most previous work, we equip models with the ability to directly access a large database to perform predictions—a semi-parametric approach.

\begin{figure}[t]
\begin{center}
\includegraphics[width=0.99\textwidth]{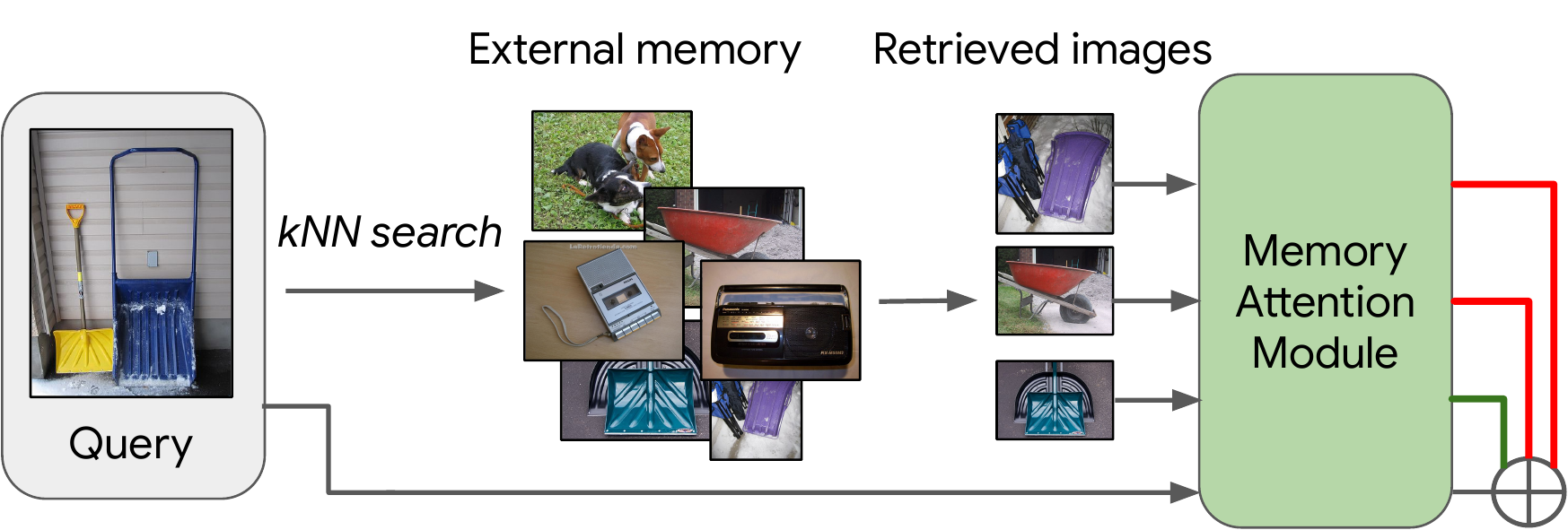} 
\end{center}
\caption{
Retrieval augmented classification finds similar images to the query from an external memory.
Our memory attention module learns the importance of each retrieved image by assigning high weights (green line in the figure) to the relevant images, and low weights (red line) to the irrelevant images.
\label{fig:teaser}
}
\end{figure}

\begin{figure*}[t]
\begin{center}
\includegraphics[width=0.99\textwidth]{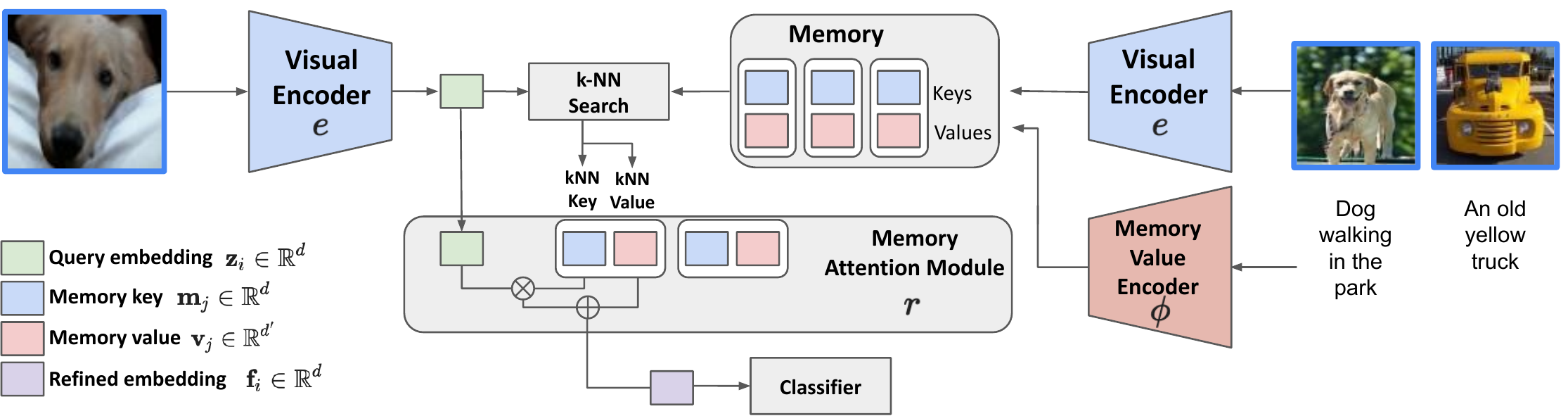}
\end{center}
\caption{
\textbf{Overview of our method.} Retrieval augmented classification aims to retrieve relevant images from an external memory dataset when making predictions. 
Each example in the memory is composed of a \emph{key} and \emph{value} embedding pair.
Key embeddings are extracted using the same visual encoder as the query image, but the value embeddings can be extracted with any other encoder. 
Both visual and value encoders are remain frozen during the training.
We perform an approximate $k$-NN search between the query embedding and memory keys to find relevant images from the memory dataset.
The retrieval module receives the query embedding and the $k$ retrieved key-value pairs from the memory.
We learn the importance of each memory example by computing the attention weights between the query embedding and the memory keys.
The memory values, weighted by their corresponding attention weights, are used to compute the refined embedding, which is then passed to the classifier.
\label{fig:overview}
}
\end{figure*}

To evaluate our approach, we focus on the problem of long-tailed recognition and learning with noisy labels. The distribution of real-world data is often noisy, imbalanced and highly skewed on a per-class basis, with a majority of classes containing a small number of samples. Long-tailed recognition is a well-studied problem~\cite{He2009knowledge, Park2008TheLT}. Base approaches are largely variants of the core idea of “adjustment”, where the learner is encouraged to focus on the tail of the distribution. This is achieved either by re-weighting samples during training~\cite{huang2016cvpr} and cluster-based sampling~\cite{cui2021parametric}, logit or loss modification~\cite{Deng_2021_CVPR, menon2021longtail, DBLP:journals/corr/abs-2104-06094, hong2021disentangling} or ensembling~\cite{wang2020long}. Despite being well-studied, commonly occurring, and of great practical importance, classification performance on long-tail distributions lags significantly behind the state of the art for better balanced classes~\cite{Guo_2021_CVPR, xiang2020learning, zhou2020BBN}.

Long et al.~\cite{long2022retrieval} introduce a retrieval-augmented classification model that explicitly stores the tail knowledge. 
In comparison to this work, we suggest to retrieve from a web-scale vision-text database and augment the input query with the retrieved knowledge using a memory attention module, before making class predictions.
We design the external memory as pairs of key-value embeddings.
These embeddings are computed by encoding vision and language data from multiple sources (Web images with alt-text such as Webli~\cite{pali2022}, LAION~\cite{schuhmann2021laion}, YFCC100M~\cite{thomee2016yfcc100m} datasets as well as image classification datasets like ImageNet~\cite{russakovsky2015imagenet}).
Memory key embeddings are used to retrieve the $k$-nearest neighbors of the input query vectors. 
Our memory attention module learns the importance of each retrieved memory example by computing attention weights between the query embedding and memory keys.
Relevant examples have more influence, whereas the contribution of the irrelevant noisy examples is down-weighted.
Learned attention weights are then used to combine memory values and produce a refined embedding, which is then used to make class predictions. 
Figure~\ref{fig:teaser} shows a high-level visualization of our method.

Our contributions are summarized as follows:
\begin{compactitem}
\item 
We propose a retrieval-augmented recognition model that explores efficient means of augmenting visual models with a massive-scale memory without significantly increasing computations.
\item
We propose a simple yet powerful way to fuse the retrieved knowledge with the input query using a memory attention module. 
\item
Our method achieves state-of-the-art results on various benchmarks, such as long-tail recognition and learning with noisy labels. We achieve $78.9$ accuracy on ImageNet-LT dataset, 50.3 accuracy on Places-LT dataset, and 83.6 on Webvision dataset.
\end{compactitem}

%% file: related.tex
\section{Related work}

External memory collections have been used for various tasks in computer vision and other domains such as NLP.
One of the earliest works combining deep network with an external memory is \emph{Neural Turing Machines}~\cite{graves2014neural}, where an external memory is updated with \emph{write} and \emph{erase} operations and a learned controller.
Santoro~\etal~\cite{santoro2016mann} propose MANN (memory-augmented neural network), where a differentiable external memory is utilized for meta-learning.

In the NLP domain, memory-based methods have been used to access external large-scale datasets.
Khandelwal~\etal~\cite{khandelwal2019generalization} propose to interpolate the outputs of a trained language model with a non-parametric $k$-NN model.
REALM~\cite{guu2020realm} retrieves external knowledge from Wikipedia for question answering.
Lewis~\etal~\cite{lewis2020retrieval} use an external memory to generate questions and answers, which are then used for question answering.
Wang~\etal~\cite{wang2022training} retrieve nearest neighbors of each training example from the training set, and combine each input with the retrieved content.
They show the benefit of this approach in various NLP tasks, such as summarization, language modeling and machine translation.
Wu~\etal~\cite{wu2022memorizing} define the memory as previously seen words in the same document, and learn how to combine them with the input tokens in a transformer.

RETRO~\cite{borgeaud2022improving} systematically evaluates the impact of large-scale external memory datasets for NLP tasks.
Our paper is similar to RETRO~\cite{borgeaud2022improving}, in that we also utilize a large-scale external memory, but in the vision domain.
Similar to RETRO, we use frozen feature extractors to reduce the complexity of large-scale $k$-NN search, in order to focus on the benefits of massive-scale external knowledge sources.

Recent methods in computer vision also make use of external memory for various tasks.
Iscen~\etal~\cite{iscen2022memory} use a memory to store previously seen examples in incremental learning.
Nakata~\etal~\cite{nakata2022revisiting} store feature maps from the training set in the memory, and perform $k$-NN for classification.
Chen~\etal~\cite{chen2022re} and Blattmann~\etal~\cite{blattmann2022semi} retrieve nearest neighbors from a memory for generative vision models.
Basu~\etal~\cite{basu2022generalization} study the generalization of retrieval-based models from a theoretical perspective.

Perhaps the most similar method to our own is \emph{Retrieval Augmented Classification} (RAC)~\cite{long2022retrieval}. 
The authors combine the output of a \emph{base} model, \ie a typical vision encoder, and the retrieval module.
 The retrieval module is learned by first finding $k$-NN of the visual input from the memory based on visual embeddings.
Then the corresponding raw text labels of the $k$-NN are concatenated, and a textual embedding is extracted with a pre-trained CLIP model~\cite{radford2021clip}.

Our work is different in that we do not assume that every retrieved example has the same importance.
By concatenating the raw text labels of each retrieved example in a single sequence of text, RAC assigns the same importance to each retrieved item.
In contrast, we explicitly learn the contribution of each retrieved item, and weight them accordingly.
Additionally, RAC uses the training set itself as the external memory.
In our work, we rigorously evaluate different candidate datasets of varying scales for the external memory.
Our experiments use memory datasets up to $1$B images, and show that larger memory datasets show benefits.

%% file: method.tex
\section{Method}
\label{sec:method}

\emph{Retrieval augmented classification} aims to enhance the query input by retrieving relevant images from an external memory.
In this section we first formulate our task, then propose different alternatives to fuse the query and the retrieved information.
Figure~\ref{fig:overview} shows an overview of our method.

\head{Problem formulation.}
Let us define a \emph{downstream} dataset of $N$ images by $X \defn \{x_1, \ldots, x_N\}$.
Our task is \emph{supervised} classification, meaning that each image is accompanied by its label $Y \defn (y_1, \ldots, y_N)$ with $y_i \in \real^C$, where $C$ is the number of classes.
In a typical classification problem, our goal is to learn a model which takes an input image $x_i$, and maps it to class prediction scores, \ie \emph{logits}. 
The model consists of two parts; the visual encoder, and the classifier.
The visual encoder $e: x \rightarrow \real^d$ takes an input image $x_i$ and maps it to a $d$-dimensional vector, \ie $\vz_i \defn e(x_i) \in \real^d$.
The resulting feature embedding $\vz_i$ is then passed to the classifier $h: \vz \rightarrow \real^C$ to obtain the logits.
The model output, \ie logits, are denoted as:
\begin{equation}
f(x_i) = h(e(x_i)) = h(\vz_i).
\label{eq:model}
\end{equation}

The model parameters are trained by minimizing any supervised loss function, such as cross-entropy, or LACE loss~\cite{menon2021longtail} when the training data is imbalanced.

\subsection{Retrieval augmented classification}
\label{sec:retrieval}

Typically, classification models are trained to make predictions only considering the images $x_i$ in the downstream dataset~\eqref{eq:model}.
The classifier $h(.)$ takes a single input $x_i$ and outputs the class logits.

Retrieval augmented classification aims to train more robust and accurate models by leveraging relevant information from an external source of knowledge, \ie a \emph{memory dataset} $M \defn \{m_1, \ldots, m_L\}$, for each downstream image $x_i$.
More specifically, the model predictions now also depend on $M$, in addition to $x_i$.
Note that $M$ is collected independently from $X$.
It is therefore not guaranteed that it will contain relevant information \wrt to the each $x_i \in X$.
We also do not assume that the memory dataset $M$ contains class labels, but the images may be accompanied by additional information, such as free-form text.

In practice, only the most relevant subset of $M$ is directly used for classification for a given training image $x_i$.
Let $\vM = [e(m_i), \dots, e(m_L)]$ be the set of feature embeddings of each image in $M$.
We compute the cosine similarity between $\vz_i$ and each embedding $\vm_j \in \vM$ to find the $k$-nearest neighbors.
The top-$k$ ranked embeddings are then used during the prediction:
\begin{equation}
f(x_i) = h(r(\vz_i, {\vM}_{\nn{}(\vz_i; \vM)})),
\label{eq:retrModel}
\end{equation}
where ${\vM}_{\nn{}(\vz_i;\vM)}$ denotes top-$k$ ranked embeddings of $\vz_i$ from $\vM$,
and $r(., .)$ is a \emph{retrieval module}, which learns how to combine $\vz_i$ with the vectors from ${\vM}_{\nn{}(\vz_i;\vM)}$.
Different choices for $r(., .)$ will be discussed in Section~\ref{sec:retrModule}.
Unlike~\eqref{eq:model}, the retrieval augmented model~\eqref{eq:retrModel} makes predictions while directly leveraging the information from $M$.

\head{Co-embedded memory.}
Long \etal~\cite{long2022retrieval} show that different types of embeddings can be extracted from the memory.
This allows us to utilize even more additional information, \eg different modalities, corresponding to the memory examples.
We will now describe this scenario in more detail.

Let us assume that there are two sets of embeddings corresponding to each example $m_i$ in the memory.
\emph{Memory keys} $\vM$, as defined above, are extracted using the same visual encoder $e(.)$ as $\vz_i$.
Let us also define \emph{memory values} as a set of vectors $\vV=[\phi(m_1), \dots, \phi(m_L)]$, extracted with an encoder $\phi: x \rightarrow \real^{d'}$.
Note that the output dimensionality of $e(.)$ and $\phi(.)$ are not necessarily equal.
As before, the $k$-NN indices are obtained by computing the cosine similarity between $\vz_i$ and $\vM$.
However, we now select the rows of $\vV$ that correspond to the indices of the $k$-NN search to make the prediction:
\begin{equation}
f(x_i) = h(r(\vz_i, {\vV}_{\nn{}(\vz_i;\vM)})),
\label{eq:coRetrModel}
\end{equation}
where ${\vV}_{\nn{}((\vz_i;\vM)}$ denotes that the $k$-NN search is done between $\vM$ and $\vz_i$, but the corresponding indices from $\vV$ are selected.

In practice, \emph{memory values} can be extracted from various different types of encoders.
One can use a larger visual encoder model, such as ViT-G/14~\cite{DBLP:conf/cvpr/Zhai0HB22}, to extract more robust visual embeddings or use a text encoder, such as T5~\cite{DBLP:journals/jmlr/RaffelSRLNMZLL20}, to take advantage of a different modality.

\subsection{Retrieval fusion module}
\label{sec:retrModule}

The retrieval module $r(. , .)$ combines the original query vector $\vz_i$ with the retrieved memory values ${\vV}_{\nn{}(\vz_i;\vM)}$.
In this section, we first introduce a simple method which is based on the mean of the retrieved memory values.
We then introduce a more powerful method, which learns the amount of contribution each memory value embedding makes to the final prediction.
For the sake of simplicity, we drop the $\vz_i$ from the notation of memory keys and values in this section, \ie ${\vM}_{\nn{}}$ and ${\vV}_{\nn{}}$ denote the $k$-NN memory keys and values of $\vz_i$, respectively.

\head{Mean $k$-NN fusion module. }
This method simply computes the mean of retrieved memory values to make the final prediction.
The refined output embedding is defined as:
\begin{equation}
r(\vz_i, {\vV}_{\nn{}}) = \vz_i + \chi\left(\frac{1}{k} \sum_{\vv \in {V}_{\nn{}}} \vv \right),
\label{eq:meanModel}
\end{equation}
where $\chi: \real^{d'} \rightarrow \real^d$ is a dense layer which maps the mean of memory value embeddings from the $d'$-dimensional vector space to the initial $d$-dimensional vector space.
We use this method as a baseline in our experiments.
It demonstrates the impact of using the retrieved memory values without learning their importance.

\head{Memory attention module (MAM). }
It is not ideal to assume that every vector in ${\vV}_{\nn{}}$ has the same importance.
Certain memory values may be more relevant for $\vz_i$, whereas others may bring noise.
We propose to compute attention weights between the query vector $\vz_i$ and the retrieved memory keys ${\vM}_{\nn{}}$, which lie in the same feature space, to learn the contribution of each vector in ${\vV}_{\nn{}}$:
\begin{equation}
r(\vz_i, {\vM}_{\nn{}}, {\vV}_{\nn{}}) = \vz_i + \chi \left(\sigma \left(\frac{\psi_Q(\vz_i) \psi_K({\vM}_{\nn{}})}{\sqrt{d}} \right) {\vV}_{\nn{}} \right),
\label{eq:attnModel}
\end{equation}
where $\sigma$ is the softmax, $\psi_Q: \real^d \rightarrow \real^d$ and $\psi_K: \real^d \rightarrow \real^d$ are dense layers, and $\chi: \real^{d'} \rightarrow \real^{d}$ is another dense layer which maps the output back to the original input space.

Note that Eq~\eqref{eq:attnModel} can be repeated $L$ times, \ie $L$ layers. 
Let
$\vf_{i}^{1} = r(\vz_i, {\vM}_{\nn{}}, {\vV}_{\nn{}})$ 
denote the output of the first layer. 
The output after $L$ layers can be computed as:
\begin{equation}
\vf_i^L = \vz_i + \chi \left(\sigma \left(\frac{\psi_Q(\vf_i^{L-1}) \psi_K({\vM}_{\nn{}})}{\sqrt{d}} \right) {\vV}_{\nn{}} \right).
\label{eq:attnDeepModel}
\end{equation}

Similar attention mechanisms are used for different purposes, such as when learning the relationship between the query vector and class vectors~\cite{tian2022vl}.
We show in our experiments that MAM is an excellent option for retrieval augmented classification as well, significantly outperforming the other baselines while achieving state-of-the-art accuracy in various tasks.

%% file: memory.tex
\section{Memory}
\label{sec:memory}

In this section, we first discuss different ways of constructing the memory dataset $M$. 
We then describe different ways of computing the embeddings for memory keys ($\vM$) and values ($\vV$).

\subsection{Memory datasets}
\label{sec:memorySets}

We use various memory datasets of different sizes in our experiments. 
We will now describe them in more detail.
Each of the choices below are thoroughly evaluated in Section~\ref{sec:expmemory}.

\head{Downstream dataset.}
This is the most straightforward choice for choosing the memory dataset. 
Under this setting, the memory set $M$ and the downstream dataset $X$ are the same.
This guarantees that there will be at least one relevant memory example $m_i$ for each $x_i$, as most datasets have multiple instances of each class.
One disadvantage of this choice is that most of the downstream datasets do not contain rich free-form text descriptions.
They have textual class labels, but those do not change between the different instances of the same class, and consequently are not very  discriminative.

\head{YFCC.}
The Yahoo Flickr Creative Commons dataset~\cite{thomee2016yfcc100m} contains approximately $100$M images.
Each image is accompanied by various metadata, including free-form text descriptions.
We use the subset of $15$M images as the recent work~\cite{radford2021clip,zhai2021lit}.
This subset is created by choosing images that have English text of high quality.

\head{LAION.}
LAION dataset~\cite{schuhmann2021laion} contains $400$M image-text pairs.
The dataset was built by gathering image-text pairs from random web pages.
The low-quality pairs are removed by computing the cosine similarity between their CLIP~\cite{radford2021clip} embeddings.
This results in $400$M image-text pairs collected from the web.

\head{WebLI.}
WebLI dataset~\cite{pali2022} contains over $10$B pairs of image-text pairs from 100 languages.
It is built from publicly available image and text data from web pages.
It contains various metadata for each image, including free-form text description.
We use the subset of $1$B images used to train the PaLI model~\cite{pali2022}.
This subset is created by scoring image-text pairs based on cross-model similarity.
Then a threshold is applied on cross-modal similarity scores, which ends up retaining about $1$B image-text pairs.

\head{All.}
This is the combination of all the memory datasets mentioned above, \ie downstream dataset, YFCC, LAION and WebLI.
This variant has approximately $1.5$B image-text pairs in the memory dataset.

\subsection{Memory keys and values}
\label{sec:memoryValues}

The parameters of the visual encoder $e(.)$, which is used to extract query vectors $\vz_i$ and memory keys $\vM$, are frozen during training.
This choice allows us to efficiently use memory datasets with up to $1$B images, as the memory keys $\vM$ are computed and indexed only once offline for efficient $k$-NN search with the query vector $\vz_i$.
A similar strategy has been employed in RETRO~\cite{borgeaud2022improving}, where the authors show the benefit of retrieval augmentation with frozen text embeddings.
We use the ViT-B/16~\cite{dosovitskiy2020image} model trained on the JFT-3B dataset~\cite{DBLP:conf/cvpr/Zhai0HB22} for the visual encoder $e(.)$, as it has shown to be a powerful vision encoder.

We explore different choices of $\phi(.)$ to compute memory values $\vV$.
Memory value encoder $\phi(.)$ is also fixed, and its parameters are frozen during training.
This allows us to use very large models for $\phi(.)$, as the memory values $\vV$ are only computed once offline, and the actual model is not needed during the training or inference.

\head{Visual encoders.}
We choose more powerful and bigger visual encoders as $\phi(.)$.
More specifically, we choose pre-trained ViT-L/16, ViT-g/14 and ViT-G/14 architectures~\cite{DBLP:conf/cvpr/Zhai0HB22} to compute $\vV$.
ViT-G/14 has $2$B parameters, making it challenging to load it in a GPU memory during the training.
However, because $\vV$ is computed offline, we only access the ViT-G/14 embeddings, not the model itself, during the training.

\head{Text encoders.}
Another way of extracting the memory values is to exploit other modalities from $M$. 
We use the pre-trained T5-Base~\cite{DBLP:journals/jmlr/RaffelSRLNMZLL20} text encoder, to extract textual embeddings as $\vV$.
If the memory set does not have free-form text descriptions, \eg if the memory set is the downstream dataset, we extract textual embeddings from the text labels of each image.
Text labels are turned into sentences by using pre-defined prompts~\cite{radford2021clip}.

\subsection{Retrieval complexity}
Our method requires a $k$-NN search between the input query vector and the memory keys.
We use the SCaNN library~\cite{guo2020accelerating} to perform approximate $k$-NN search in our experiments.
It has a sublinear complexity, meaning that it takes $\mathcal{O}(\log{}M)$ for a memory dataset of $M$ elements.
In practice, querying a memory dataset of $1$B elements takes only milli-seconds.
Because we use a fixed vision encoder in our experiments, we pre-compute and save all $k$-NN, which speeds up training time.

%% file: experiments.tex
\begin{figure*}
\input{fig_ablation_single}
\caption{\textbf{Ablation study on ImageNet-LT.} \textbf{Left}: We show the impact of different memory sets and memory value encoders. We set $k=100$ for this experiment. \textbf{Right}: We show the impact of $k$ for different memory sets. We use T5-Base to represent memory values for this experiment.  
\label{fig:ablation} 
}
\end{figure*}
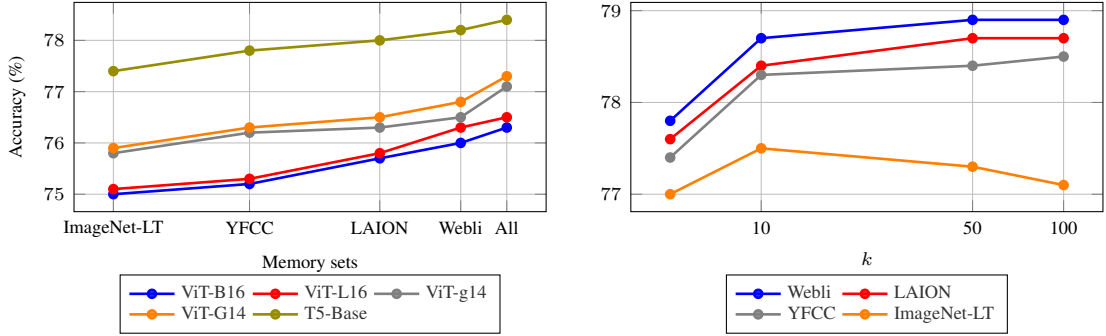

\section{Experiments}

In this section, we first describe the downstream datasets used in our experiments and detail our experimental setup.
We then conduct a rigorous study showing the impact of different memory datasets and memory value encoders.
Finally, we compare our results against the existing methods in the literature.

\subsection{Experimental setup}
\label{sec:setup}

We report experiments for three different image classification tasks: long-tailed recognition, learning with noisy labels, and fine-grained classification. 
We now describe the downstream datasets that we use for each of these tasks.

\head{Long-tailed Recognition.}
Long-tailed recognition assumes that there is a strong imbalance in terms of images per class in the training set.
The goal is to learn robust classifiers that can accurately classify every class, regardless of the number of times it appears in the training set.

We use two datasets for this task: ImageNet-LT~\cite{OLTR} and Places-LT~\cite{OLTR}.
ImageNet-LT has $1000$ classes and the number of training images per class varies from $5$ to $1,280$.
It is created by taking a subset of the original ImageNet dataset~\cite{ImageNet}, so that the number of images per class follows a long-tailed distribution.

Similarly, Places-LT is created by taking a subset of the original Places365 dataset~\cite{zhou2017places}, such that the number of training images per class follows a long-tailed distribution.
There are $365$ classes in this dataset, and the number of images per class varies from $5$ to $4,980$.

The validation sets for both datasets are balanced.
We report the top-1 overall accuracy for both datasets. 
We also report the accuracy for \emph{many-shot} classes (more than 100 images per class), \emph{mid-shot} classes (between 20 and 100 images) and \emph{few-shot} classes (less than 20 images), separately, following the protocol in~\cite{OLTR}.

\head{Learning with noisy labels.}
The Webvision dataset~\cite{li2017webvision} contains $2.4$M images and $1000$ classes.
The data is collected from the web, and the labels are assigned without human supervision.
Therefore, some of the labels are noisy.
Training set is imbalanced, meaning that there are different number of examples for each class.
We report the top-1 overall accuracy for the validation set.

\head{Fine-grained classification.}
We use the iNaturalist2021-Mini dataset~\cite{van2021benchmarking} for this task.
This dataset contains fine-grained images of species, \eg insects, plants, birds \etc 
There are $10,000$ classes and $50$ images per class, making it a total number of $500,000$ training images.
The validation set contains $100,000$ images.
We report the top-1 overall accuracy.

\head{Implementation details.}
We use a frozen ViT-B/16 as the visual encoder $e(.)$ to represent the query vectors from the downstream datasets and memory key embeddings.
Unless otherwise specified, training lasts $10$ epochs, with a learning rate of $0.001$ and batch size of $512$.
The learning rate follows a warm-up schedule of $1$ epoch, and then is reduced in each epoch using a cosine decay schedule~\cite{LH16}.
We use an Adam optimizer~\cite{kingma2017adam} with a weight decay of $0.2$.
We also use label smoothing~\cite{szegedy2016rethinking} during training to prevent over-fitting.
For the Memory Attention Module (MAM)~\eqref{eq:attnModel}, we use $8$ layers, \ie $L=8$.
We retrieve $k=100$ examples from the memory, unless otherwise specified.

\subsection{Impact of the memory}
\label{sec:expmemory}

We study the effect of different choices for the memory construction in this section.
Section~\ref{sec:memory} introduces various memory datasets and memory value encoders in detail. 
We now investigate how different memory datasets and value encoders behave in the ImageNet-LT downstream dataset.

Figure~\ref{fig:ablation} (left) shows the impact of different memory value encoders in different memory datasets. 
We set $k=100$ for this experiment.
We see that the textual memory value encoder (T5-Base) obtains a better accuracy compared to visual memory value encoders.
Even much larger visual memory value encoders, such as ViT-G/14, does not outperform a smaller T5-Base model.
We believe that this behavior is due to the fact that T5-Base represents a different modality (text), which otherwise is not available to the input.
Textual memory values are thus complementary to the visual signals, which are extracted from the input query in any case.

Figure~\ref{fig:ablation} (left) also shows that the accuracy generally improves as the size of the memory dataset becomes larger. 
Larger relative improvements are especially observed for visual memory value encoders as the size of the memory dataset increases.
The accuracy for visual memory value encoders continues to increase even when all the memory datasets are combined ($\approx1.5$B examples). 

Figure~\ref{fig:ablation} (right) shows the impact of $k$ for different memory sets.
This hyperparameter controls the number of keys and values retrieved from the memory dataset.
We use textual memory value embeddings (T5-Base) for this experiment.
We see that the large memory datasets, \eg YFCC, LAION, Webli, are less sensitive to the choice of $k$, whereas the accuracy starts to decrease for a smaller memory dataset such as ImageNet-LT.
That is because ImageNet-LT only has a few positive examples for certain images.
As $k$ becomes larger, the retrieved set of vectors mostly contain noise.
That is not the case for larger memory datasets, as they are likely to have many relevant examples for each image.
Thus, they are less sensitive to larger $k$.

\subsection{Comparison to baselines}

\begin{table*}
  \input{baseline_table}

  \caption{\textbf{Comprehensive evaluation on ImageNet-LT and Places-LT.} The accuracy for many-shot ($>100$ images), mid-shot ($20$-$100$ images) and few-shot ($<20$ images) classes are reported separately. \textbf{Top:} We report the results for various baselines. \textbf{Bottom: } We compare our method against the existing methods in the literature. RAC$\dagger$ denotes our re-implementation of RAC~\cite{long2022retrieval} in exactly the same setting as our method. \faLock ~means that the visual encoder is frozen during the downstream task training.
  \label{tab:baseline}}
\end{table*}

In this section, we show the benefit of our Memory Attention Module (MAM) by comparing it with different baselines. 
We report the accuracy for the following baselines.
\emph{Linear} classifier learns a fully connected layer on top of frozen downstream dataset embeddings.
\emph{MLP} classifier is a two-layer MLP with non-linearity in between the two layers.
Linear and MLP classifiers do not use an external memory dataset for retrieval. 
\emph{Mean $k$-NN} computes the mean of the retrieved memory values; it do not learn the contribution of each retrieved memory value.
See Section~\ref{sec:method} for more details.
We use the WebLI memory dataset and T5-Base memory values for this experiment.

Table~\ref{tab:baseline} (Top) shows the accuracy for many-shot, mid-shot, and low-shot classes separately for all the baselines on ImageNet-LT and Places-LT datasets.
All the methods are trained with the LACE~\cite{menon2021longtail} loss, which has a balancing effect between the low-shot and many-shot classes.
Nevertheless, the low-shot accuracy suffers for the methods that do not use retrieval, \eg linear and MLP classifiers.

On the other hand, retrieval-based methods, \eg mean $k$-NN and MAM (Ours), are less prone to over-fitting on many-shot classes.
However, mean $k$-NN overcompensates for the low-shot classes by sacrificing the accuracy for the many-shot classes on ImageNet-LT.
This is not the case for our method, which achieves the highest overall accuracy by performing well across all three class categories.
Similar observations can be made in the Places-LT dataset. 
Our method achieves the highest accuracy on many-shot, mid-shot and low-shot classes, and the highest accuracy overall.

The experiments on Table~\ref{tab:baseline} (Top) demonstrate that the retrieval augmented classification alone does not always improve the classification accuracy.
This is evidenced by the performance of \emph{mean $k$-NN}.
On the other hand, as we learn the contribution of the each retrieved example from the memory, we are able to filter out the noisy examples more accurately.
This results in the highest accuracy overall, while not sacrificing the accuracy for the many-shot, mid-shot and low-shot classes.

\begin{figure*}
\begin{center}
\includegraphics[width=0.8\textwidth]{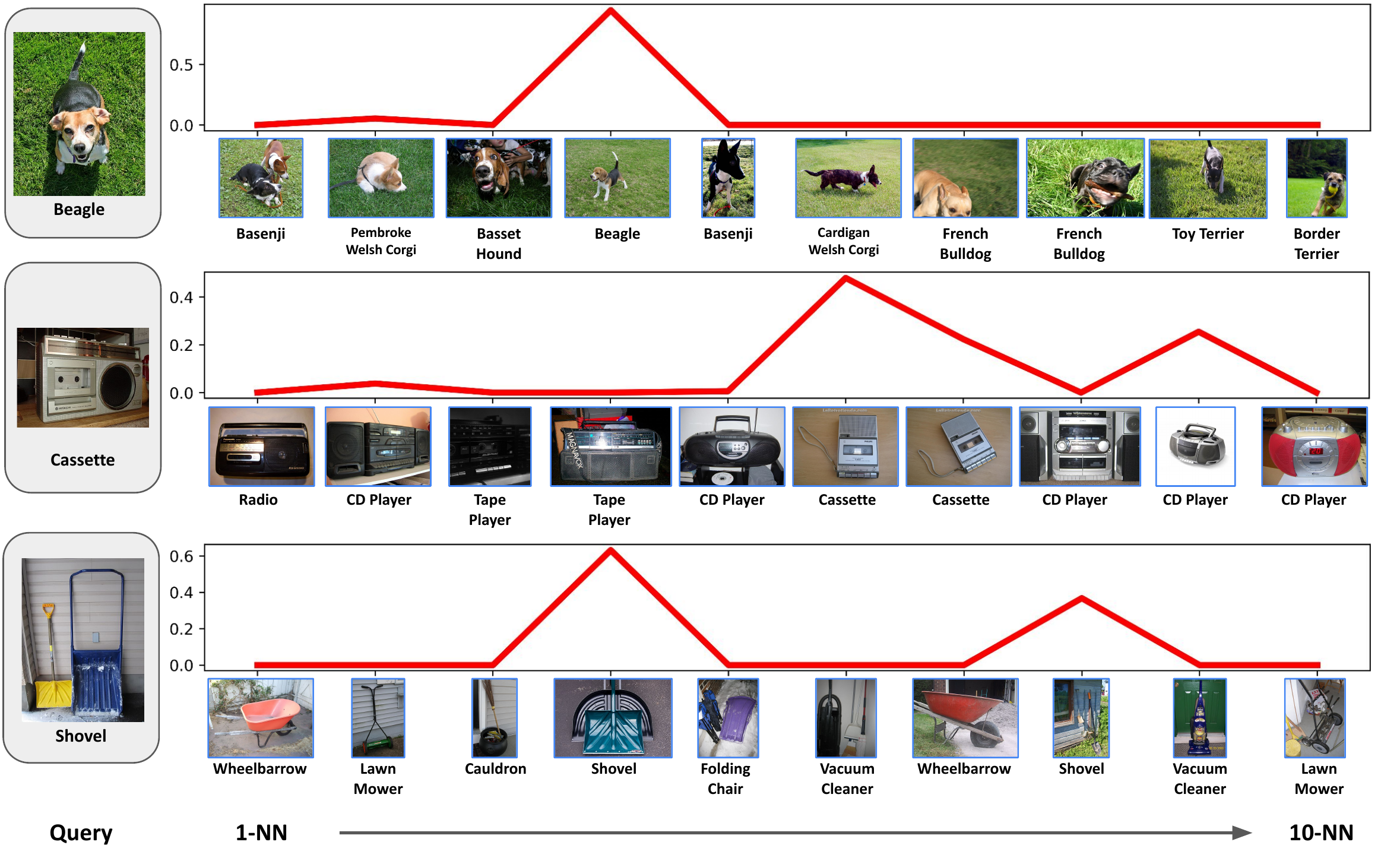}
\end{center}
\caption{
\textbf{Qualitative Examples.} 
We present the output of our method visually. 
We conduct this experiment by choosing the ImageNet-LT dataset as the query and memory dataset.
We display the query images from the test set on the left.
Their $k$-NN from the training set are displayed on the right, and ordered from left to right.
We display the attention weight assigned to each $k$-NN above the corresponding image.
\label{fig:qual}
\vspace{-0.2cm}
}
\end{figure*}

Table~\ref{tab:other} shows the comparison of our method against the baselines for fine-grained classification (iNaturalist2021-Mini) and learning with noisy labels (Webvision). 
We use WebLI as the memory dataset, and T5-Base as the memory value encoder for these experiments.
We observe that our method displays consistent improvement in both datasets.
Note that it overperforms the state-of-the-art in Webvision, without finetuning the visual encoder like the existing work.
This shows that our method is suitable for various classification tasks, and not only long-tailed recognition.

\begin{table}
  \input{other_table}
  \caption{\textbf{Evaluation on iNaturalist2021-Mini and Webvision.} We compare our method against the baselines and existing work in iNaturalist2021-Mini (fine-grained classification) and Webvision (learning with noisy labels) downstream datasets.
  \label{tab:other}}
\end{table}

\subsection{Comparison to existing methods}

We now compare our method against the state-of-the-art.
Table~\ref{tab:baseline} (Bottom) shows the accuracy of the prior art in ImageNet-LT and Places-LT datasets.
VL-LTR~\cite{tian2022vl} and RAC~\cite{long2022retrieval} use the same ViT-B/16 backbone as our method.
However the pre-training of the ViT-B/16 differs between different methods.
VL-LTR uses the ViT-B/16 pre-trained with CLIP~\cite{radford2021clip}, whereas RAC uses an ImageNet-21k pre-trained ViT-B/16 architecture~\cite{dosovitskiy2020image}.
Both methods finetune the visual encoder on the downstream dataset.
In this paper, we use a ViT-B/16 visual encoder pre-trained on the JFT-3B dataset~\cite{DBLP:conf/cvpr/Zhai0HB22}.
We also re-implement RAC with our visual and text encoder (T5-Base) for a better comparison in Table~\ref{tab:baseline}, and denote this variant as RAC$\dagger$.

Table~\ref{tab:baseline} shows that the VL-LTR achieves the highest many-shot accuracy on both datasets.
Nevertheless, this comes at the expense of a poor performance for low-shot classes.
RAC, an existing retrieval augmented classification method, shows a more balanced performance between many, mid and low-shot classes.
Our method achieves the highest accuracy on both datasets by obtaining high accuracy across different categories. 
For example, we do not achieve the highest many-shot nor low-shot accuracy on ImageNet-LT.
However, because we do not favor any category of classes above others, we have higher performance across different categories and achieve the highest overall accuracy.

\head{Fine-tuning the visual encoder.} 
In Table~\ref{tab:baseline}, we also include a variant of our method which fine-tunes the visual encoder $e(.)$ while learning the memory attention module.
We denote this variant by \emph{Ours + FT}.
The k-NN search is still done with a pre-trained, frozen vision encoder as in previous experiments.
We also include a variant of RAC$\dagger$ which follows this setup in Table~\ref{tab:baseline}.
Our method achieves even further gains of accuracy when fine-tuning the vision encoder along with the memory attention module.

\subsection{Qualitative examples}

We present some of the qualitative examples in Figure~\ref{fig:qual}.
We use ImageNet-LT as both the downstream task and the memory dataset for this task.
We display the query images on the left, and the top-$10$ retrieved nearest neighbors on the right.
The retrieved images are ordered such that left-most image is the closest one.
Above each retrieved image, we display the learned attention value of our method. 

We see that our method assigns higher attention weights to the semantically correct images from the $k$-NN list.
We observe this even if there is only one correct example in the k-NN list (\eg the \emph{beagle} query).
When there are multiple relevant images, all relevant examples get higher attention weights (\eg \emph{cassette} and \emph{shovel} queries).
The original rank position does not matter much for our method.
For example, one of the relevant retrieved images for the \emph{shovel} query has originally rank eight, but receives the second highest attention weight from our method.
Figure~\ref{fig:plqual} in Appendix shows the qualitative examples in Places-LT dataset.

%% file: fig_ablation_single.tex
\centering
\input{fig/data/sample}
\begin{tabular}{cc}
{
\begin{tikzpicture}
    \tikzstyle{every node}=[font=\scriptsize]
\begin{axis}[%
  width=0.45\textwidth,
  height=0.25\textwidth,
  xlabel={Memory sets},
    xmode=log,
    grid=both,
  ylabel= {Accuracy (\%)},
  xminorticks=false,
  xtick={1, 15, 200, 1000, 2500},
  xticklabels={ImageNet-LT, YFCC, LAION, Webli, All},  
  legend cell align={left},
  legend pos=outer north east,
  legend style={at={(0.5,-0.3)},anchor=north, font =\scriptsize, fill opacity=0.8, row sep=-2.5pt, legend columns=3},
]

  \addplot[color=blue,     solid, mark=*,  mark size=1.5, line width=1.0] table[x=scale, y expr={\thisrow{b16}}] \singlelayermemvals; \leg{ViT-B16}
  \addplot[color=red,     solid, mark=*,  mark size=1.5, line width=1.0] table[x=scale, y expr={\thisrow{l16}}] \singlelayermemvals; \leg{ViT-L16}
  \addplot[color=gray,     solid, mark=*,  mark size=1.5, line width=1.0] table[x=scale, y expr={\thisrow{g14}}] \singlelayermemvals; \leg{ViT-g14}
  \addplot[color=orange,     solid, mark=*,  mark size=1.5, line width=1.0] table[x=scale, y expr={\thisrow{bigg14}}] \singlelayermemvals; \leg{ViT-G14}
  \addplot[color=olive,     solid, mark=*,  mark size=1.5, line width=1.0] table[x=scale, y expr={\thisrow{t5base}}] \singlelayermemvals; \leg{T5-Base}

\end{axis}
\end{tikzpicture}
}

&

{
\begin{tikzpicture}
    \tikzstyle{every node}=[font=\scriptsize]
\begin{axis}[%
  width=0.45\textwidth,
  height=0.25\textwidth,
  xlabel={$k$},
    xmode=log,
    grid=both,
  xminorticks=false,
  ytick={77, 78, 79, 80},
  xtick={1, 10, 50, 100},
  xticklabels={1, 10, 50, 100},  
  legend cell align={left},
  legend pos=outer north east,
  legend style={at={(0.5,-0.3)},anchor=north, font =\scriptsize, fill opacity=0.8, row sep=-2.5pt, legend columns=2},
]

  \addplot[color=blue,     solid, mark=*,  mark size=1.5, line width=1.0] table[x=k, y expr={\thisrow{webli}}] \k; \leg{Webli}
  \addplot[color=red,     solid, mark=*,  mark size=1.5, line width=1.0] table[x=k, y expr={\thisrow{laion}}] \k; \leg{LAION}
  \addplot[color=gray,     solid, mark=*,  mark size=1.5, line width=1.0] table[x=k, y expr={\thisrow{yfcc}}] \k; \leg{YFCC}
  \addplot[color=orange,     solid, mark=*,  mark size=1.5, line width=1.0] table[x=k, y expr={\thisrow{imnet}}] \k; \leg{ImageNet-LT}

\end{axis}
\end{tikzpicture}
}

\end{tabular}

%% file: baseline_table.tex
\small
\setlength{\tabcolsep}{4pt}
\begin{center}
\begin{tabular}{lccccccccccc}
\toprule
					& 				&				& \multicolumn{4}{c}{ImageNet-LT} & \multicolumn{4}{c}{Places-LT} \\
\midrule 
Method 				&	Retrieval	& Backbone	& Many-shot 	& Mid-shot 		& Low-shot 	& All 	& Many-shot 	& Mid-shot 		& Low-shot 	& All				\\
\midrule
									 \multicolumn{11}{c}{\textsc{\textbf{Baselines}}}																																														\\
\midrule
Linear Classifier 	& 				& ViT-B16 \faLock		& 76.5 			& 72.6 					& 66.5 						& 73.5  				& 44.5							& 44.4 					& 44.0 						& 44.3  			\\
MLP Classifier 		&				& ViT-B16 \faLock		& 80.1 			& 74.1 					& 66.9 						& 75.2  				& 48.6							& 46.1 					& 41.3 						& 46.0  			\\
Mean $k$-NN  		& \checkmark	& ViT-B16 \faLock		& 75.9 			& 75.8 					& \textbf{75.7} 			& 75.8  				& 44.3							& 45.2 					& 45.5 						& 44.9  			\\
\midrule
									 \multicolumn{11}{c}{\textsc{\textbf{Existing methods}}}																																							\\
\midrule
PaCo~\cite{cui2021parametric} 		  & 				& ResNext-101	& 68.2 			& 58.7 			& 41.0 		& 60.0  	& 36.1			& 47.9 			& 35.3 		& 41.2  			\\

VL-LTR~\cite{tian2022vl}		  &					& ViT-B16				& 84.5 			& 74.6 			& 59.3 		& 77.2  		& \textbf{54.2}		& 48.5 			& 42.0 		& 50.1  			\\
RAC~\cite{long2022retrieval}  		  & \checkmark	& ViT-B16				& - 			& - 			& - 		& -  			& 48.7				& 48.3 			& 41.8 		& 47.2  			\\
RAC$\dagger$~\cite{long2022retrieval} & \checkmark	& ViT-B16 \faLock		& 80.9 			& 76.0 			& 67.5 		& 76.7  		& 50.3				& 48.3 			& 42.5 		& 47.9  				\\
RAC$\dagger$~\cite{long2022retrieval} & \checkmark	& ViT-B16 				& \textbf{85.9} & 79.3 			& 69.3 		& 80.5  		& 51.9				& 49.8 			& 46.8 		& 50.0  				\\
\midrule
\textbf{Ours} 		  		& \checkmark		& ViT-B16 \faLock		& 80.6  		& 77.5 			& 74.5 			& 78.3 			& 50.9	& 49.9 				& 47.5				& 49.9  	\\
\textbf{Ours + FT} 		  	& \checkmark		& ViT-B16				& 85.4	& \textbf{81.5} & \textbf{76.4} & \textbf{82.3} & 52.4	& \textbf{52.0} 	& \textbf{48.5} 	& \textbf{51.4}  	\\
\bottomrule
\end{tabular}
\end{center}

%% file: other_table.tex
\small{
    \scalebox{0.99}{  
        \centering
        \begin{tabular}{lcc}
        \toprule
                                                                & iNat2021-Mini    & WebVision     \\
        \midrule
        \multicolumn{3}{c}{\textsc{\textbf{Baselines}}}                                                    \\
        \midrule
        Linear Classifier                                       & 58.8              & 78.1                 \\ 
        MLP Classifier                                          & 59.6              & 81.0                  \\
        Mean $k$-NN                                             & 58.9              & 78.2                  \\
        \midrule
        \multicolumn{3}{c}{\textsc{\textbf{Existing methods}}}                                              \\
        \midrule
        MILe~\cite{rajeswar2021multi}                           & --                & 75.2                  \\
        Heteroscedastic~\cite{collier2021correlated}            & --                & 76.6                  \\
        NCR~\cite{iscen2022learning}                            & --                & 76.8                  \\
        CurrNet~\cite{guo2018curriculumnet}                     & --                & 79.3                 \\
        \midrule
        \textbf{Ours}                                           & \textbf{66.2}     & \textbf{83.6}         \\
        \bottomrule
        \end{tabular}
    }
}

%% file: conclusions.tex
\section{Conclusions}
We propose a simple, yet effective memory attention module for retrieval augmented classification in this work.
Our method learns the importance of each retrieved example, and weights their contributions accordingly.
We also present a systematic study of different memory designs, showing the benefit of massive-scale memory datasets up to $1$B image-text pairs.
The effectiveness of our method is shown by the fact that it achieves state-of-the-art results in long-tailed recognition, learning with noisy labels and fine-grained classification tasks.

%% file: appendix.tex
\twocolumn[{%
\renewcommand\twocolumn[1][]{#1}%
 \centering
 \Large
 \textbf{Improving Image Recognition by Retrieving from Web-Scale Image-Text Data} \\
 \vspace{0.5em}Supplementary Material \\
 \vspace{1.0em}
 \author{
Ahmet Iscen \ \ \ \ Alireza Fathi\ \ \ \ Cordelia Schmid\\
{\fontsize{11}{13}\selectfont Google Research}\\
}
\pagestyle{plain}
\begin{center}
\includegraphics[width=0.9\textwidth]{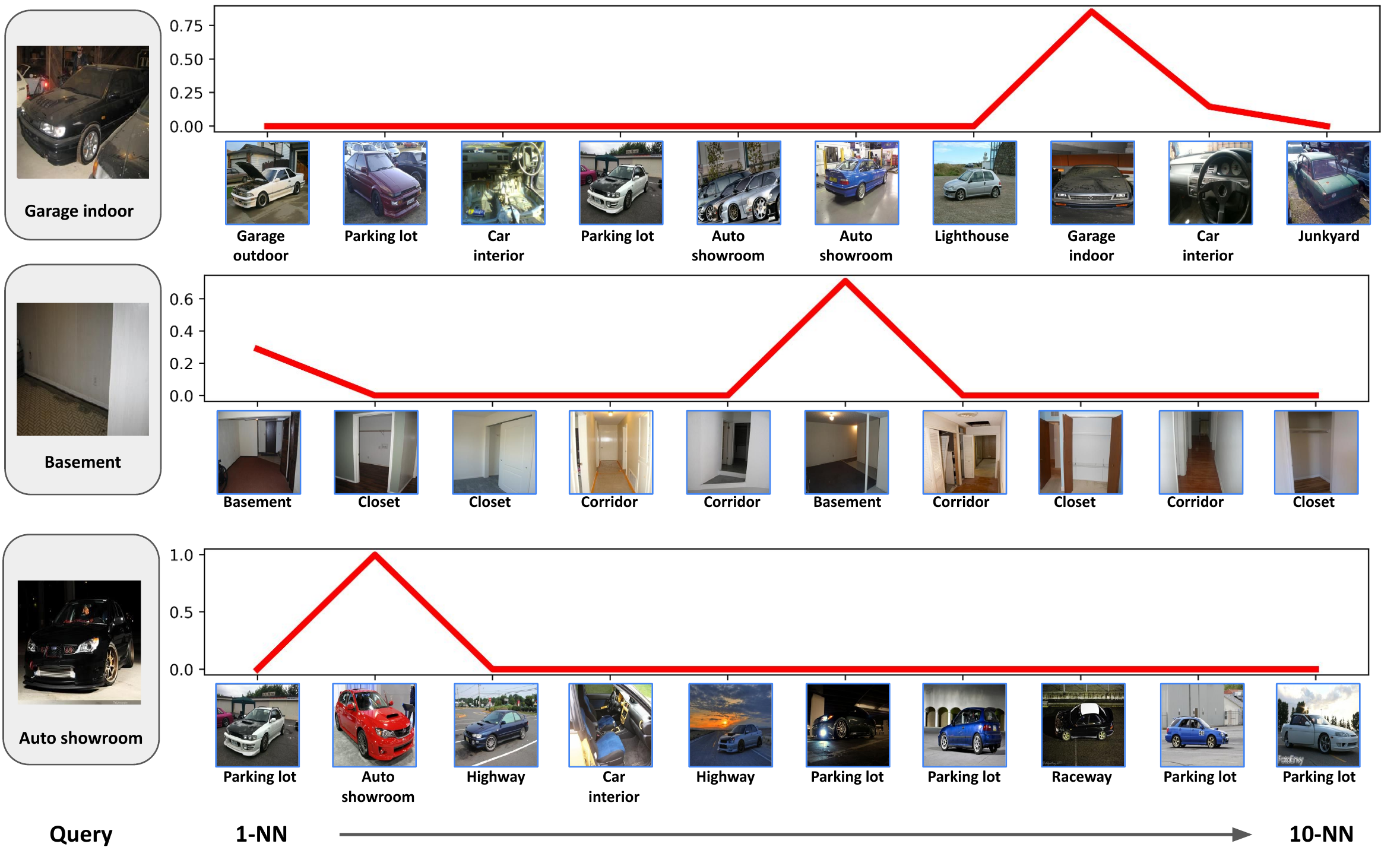}
\captionof{figure}{\small 
\textbf{Qualitative Examples.} 
We present the output of our method visually. 
We conduct this experiment by choosing the Places-LT dataset as the query and memory dataset.
We display the query images from the test set on the left.
Their $k$-NN images from the memory set are displayed on the right, and ordered from left to right.
We display the attention weight assigned to each $k$-NN above the corresponding image.
}
\label{fig:plqual}
\end{center}
}]

\begin{figure*}[h!]
\begin{center}
\includegraphics[width=0.9\textwidth]{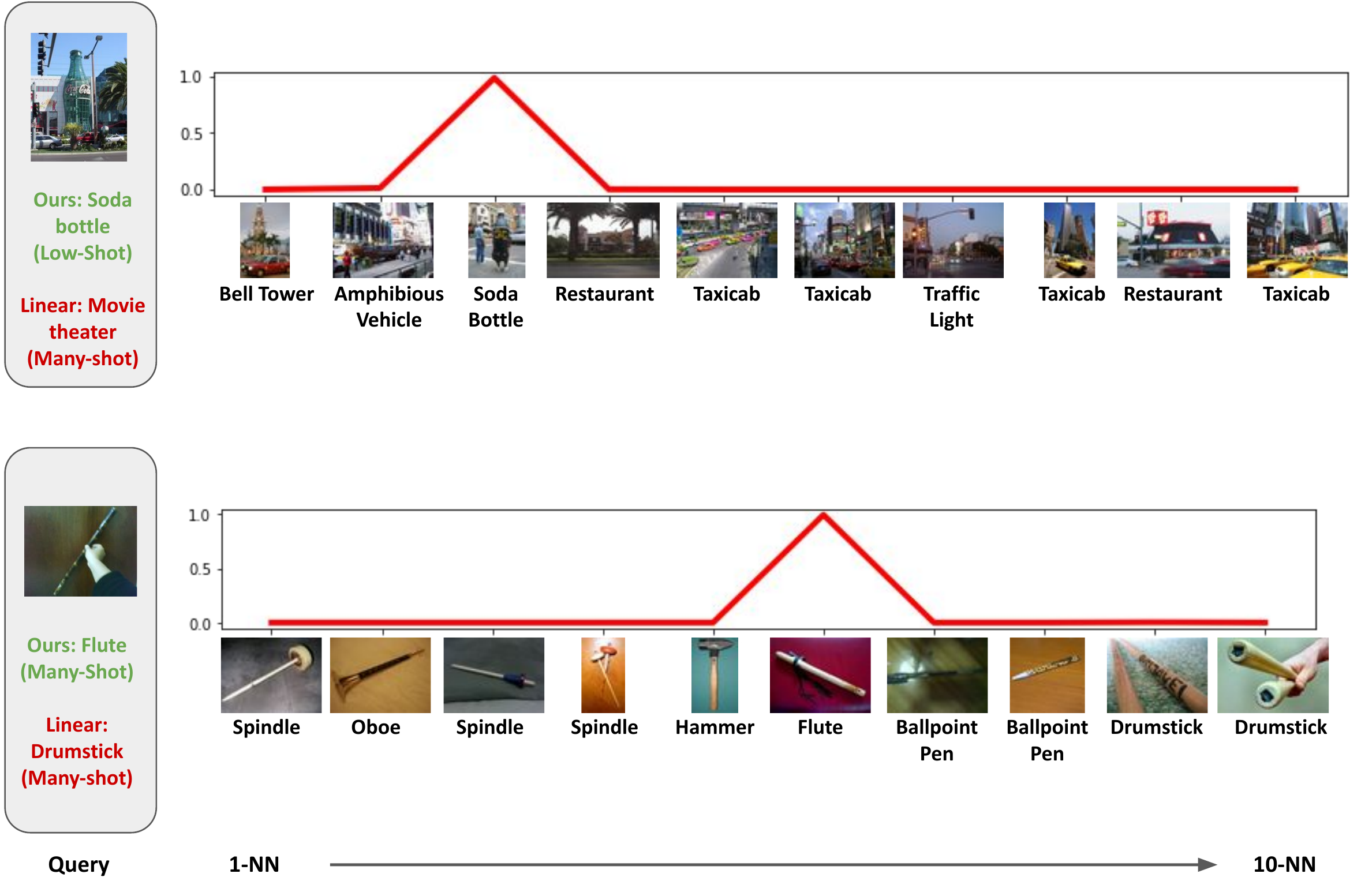}
\end{center}
\caption{
\textbf{Visual comparison on ImageNet-LT.} 
We present a visual comparison of our method and \emph{Linear}.
Below each query on the left, the correct prediction of our method is displayed in green, and the incorrect prediction of \emph{Linear} is displayed in red.
The $k$-NN images from the memory set are displayed on the right, and ordered from left to right, and their corresponding attention weights are displayed above them.
\label{fig:compqual}
}
\end{figure*}

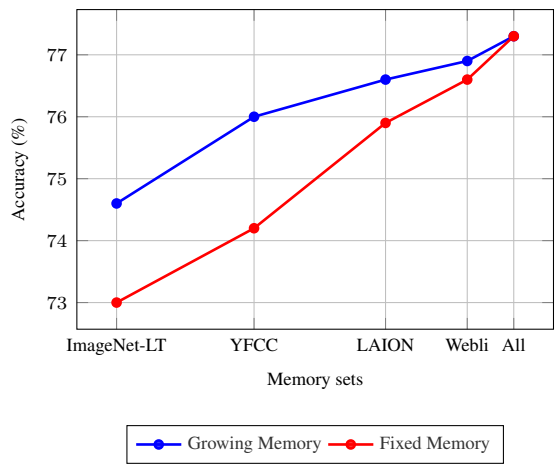
\begin{figure}
\input{fig_grow}
\caption{\textbf{Impact of growing memory on ImageNet-LT.} We train our method using the memory set denoted in the x-axis. For \emph{growing memory}, additional data is added to the memory after the training, and the inference is done with the \emph{All} memory. For \emph{fixed memory}, training and the inference is done with the same memory.
\label{fig:grow} 
}
\end{figure}

\appendix

\setcounter{page}{1}

\section{Qualitative examples for Places-LT}
Similar to Figure~\ref{fig:qual}, we present some of the qualitative examples of the Places-LT dataset in Figure~\ref{fig:plqual}.
We use Places-LT dataset as both the downstream task and the memory dataset for this experiment.
We display the query images on the left, and the top-$10$ $k$-NN on the right.

We observe that our method differentiates between different environments containing \emph{car} images.
For example, for the \emph{garage indoor} query, our method assigns a higher attention weight to the only other car image which is taken in a \emph{garage indoor}, and filters out other images of cars taken in \emph{garage outdoor}, \emph{parking lot} \etc.
We see a similar behavior for the \emph{auto showroom} query, where the images of cars taken in \emph{parking lots} are filtered out, and the only other image of a car in an \emph{auto showroom} receives a high attention weight.

For the \emph{basement} query, we see that two other \emph{basement} images receive higher attention weights.
In this case, attention weights also explain how the prediction is made.
The \emph{basement} image which is more visually similar to the query receives a higher attention weight.
This demonstrates the potential of \emph{explainability} of our method, from which we can derive how the decisions are made.

\section{Qualitative comparison with Linear}

We compare our method against the \emph{Linear} baseline qualitatively on Table~\ref{tab:other}.
We now present qualitative experiments for this comparison on Figure~\ref{fig:compqual}.
For each query on the left, we display our correct prediction in green, and incorrect \emph{Linear} prediction in red.
We also display which category of classes, \ie low-shot, mid-shot, many-shot, each of these queries belong to.

The first query is a challenging example of a \emph{soda bottle}, which is much bigger than a typical \emph{soda bottle}.
Thus, the linear classifier incorrectly assigns a more frequently seen \emph{movie theater}, which is a many-shot class, as its prediction, due to the building next to it.
On the other hand, our method correctly predicts the \emph{soda bottle} class, by retrieving another normal-than-usual soda bottle image in its $k$-NN list.

The second example, \emph{flute}, belongs to a many-shot class.
It is a visually challenging query, as it is similar to other objects such as \emph{spindle} and \emph{drumstick}.
Our method predicts the correct class by retrieving another \emph{flute} instance, whereas $k$-NN and \emph{Linear} incorrectly predict \emph{spindle} and \emph{drumstick}, respectively.

\section{Growing memory}

We now present a different scenario, where the size of the memory dataset grows during the inference.
More specifically, we train our method with either ImageNet-LT, YFCC, LAION or Webli as the memory set.
After the training, we add more image-text pairs to our memory set and it becomes \emph{All}.
The memory attention module is not re-trained after adding additional data to the memory.
Figure~\ref{fig:grow} shows that we can achieve higher accuracy without any extra training cost, by just adding image-text pairs to the memory during the inference.
The gains are more significant if our method is trained with a small memory set, \eg ImageNet-LT, but the inference is done with much larger memory set, \ie All.

%% file: fig_grow.tex
\centering
\input{fig/data/sample}
\begin{tabular}{c}
{
\begin{tikzpicture}
    \tikzstyle{every node}=[font=\scriptsize]
\begin{axis}[%
  width=0.95\textwidth,
  height=0.7\textwidth,
  xlabel={Memory sets},
    xmode=log,
    grid=both,
  ylabel= {Accuracy (\%)},
  xminorticks=false,
  xtick={1, 15, 200, 1000, 2500},
  xticklabels={ImageNet-LT, YFCC, LAION, Webli, All},  
  legend cell align={left},
  legend pos=outer north east,
  legend style={at={(0.5,-0.3)},anchor=north, font =\scriptsize, fill opacity=0.8, row sep=-2.5pt, legend columns=3},
]

  \addplot[color=blue,     solid, mark=*,  mark size=1.5, line width=1.0] table[x=scale, y expr={\thisrow{growing}}] \growmem; \leg{Growing Memory}
  \addplot[color=red,     solid, mark=*,  mark size=1.5, line width=1.0] table[x=scale, y expr={\thisrow{fixed}}] \growmem; \leg{Fixed Memory}

\end{axis}
\end{tikzpicture}
}

\end{tabular}